\documentclass[10pt,twocolumn,letterpaper]{article}

\usepackage{titling}
\usepackage{wacv}              

\usepackage{graphicx}
\usepackage{amsmath}
\usepackage{amssymb}
\usepackage{booktabs}

\usepackage{multirow}
\usepackage[accsupp]{axessibility}
\graphicspath{{images}}

\usepackage[pagebackref,breaklinks,colorlinks]{hyperref}

\usepackage[capitalize]{cleveref}
\crefname{section}{Sec.}{Secs.}
\Crefname{section}{Section}{Sections}
\Crefname{table}{Table}{Tables}
\crefname{table}{Tab.}{Tabs.}

\def\wacvPaperID{701} 
\def\confName{WACV}
\def\confYear{2024}

\begin{document}
\definecolor{Green}{RGB}{0, 170, 0}

\title{Improving Normalization with the James-Stein Estimator}

\author{Seyedalireza Khoshsirat \qquad Chandra Kambhamettu\\
Video/Image Modeling and Synthesis (VIMS) Lab, University of Delaware\\
{\tt\small \{alireza, chandrak\}@udel.edu}
}

\maketitle

\begin{abstract}
Stein's paradox holds considerable sway in high-dimensional statistics, highlighting that the sample mean, traditionally considered the de facto estimator, might not be the most efficacious in higher dimensions.
To address this, the James-Stein estimator proposes an enhancement by steering the sample means toward a more centralized mean vector.
In this paper, first, we establish that normalization layers in deep learning use inadmissible estimators for mean and variance.
Next, we introduce a novel method to employ the James-Stein estimator to improve the estimation of mean and variance within normalization layers.
We evaluate our method on different computer vision tasks: image classification, semantic segmentation, and 3D object classification.
Through these evaluations, it is evident that our improved normalization layers consistently yield superior accuracy across all tasks without extra computational burden.
Moreover, recognizing that a plethora of shrinkage estimators surpass the traditional estimator in performance, we study two other prominent shrinkage estimators: Ridge and LASSO.
Additionally, we provide visual representations to intuitively demonstrate the impact of shrinkage on the estimated layer statistics.
Finally, we study the effect of regularization and batch size on our modified batch normalization.
The studies show that our method is less sensitive to batch size and regularization, improving accuracy under various setups.
\end{abstract}

\section{Introduction}
Deep neural networks have an influential role in many applications, especially computer vision.
One milestone improvement in these networks was the addition of normalization layers \cite{krizhevsky2017imagenet,ioffe2015batch,maserat201743}.
Since then, a plethora of research has tried to improve normalization layers through various means, including better estimating the layer statistics.
In this paper, we take a statistical approach to estimate layer statistics and introduce an improved way of estimating the mean and variance in normalization layers.
The rest of this section is devoted to introducing prerequisite statistical concepts and normalization layers.
\par
\textbf{Estimators.} An estimator refers to a method used to calculate an approximation of a specific value from the data observed.
Therefore, one should differentiate between the estimator itself, the targeted value of interest, and the resulting approximation \cite{mosteller1987data}.
\par
\textbf{Shrinkage.} In statistics, shrinkage is the reduction of the effects of sampling noise.
In regression analysis, a fitted relationship appears to perform worse on a new dataset than on the dataset used for fitting \cite{everitt2010cambridge,wikipedia_shrinkage}.
More specifically, the value of the coefficient of determination `shrinks.'
\par
\textbf{Shrinkage Estimators.} A shrinkage estimator is an estimator that explicitly or implicitly uses the effects of shrinkage.
In loose terms, a naive or basic estimate is improved by combining it with other information.
This term refers to the notion that the improved estimate is pushed closer to the value provided by the `other information' than the basic estimate \cite{wikipedia_shrinkage,khoshsirat2022semantic}.
In this sense, shrinkage is used to regularize the estimation process.
In terms of mean squared error (MSE), many standard estimators can be improved by shrinking them towards zero or some other value.
In other words, the improvement in the estimate from the corresponding reduction in the width of the confidence interval is likely to outweigh the worsening of the estimate instructed by biasing the estimate towards zero \cite{james1992estimation}.

\begin{figure*}[t]
\includegraphics[width=\linewidth]{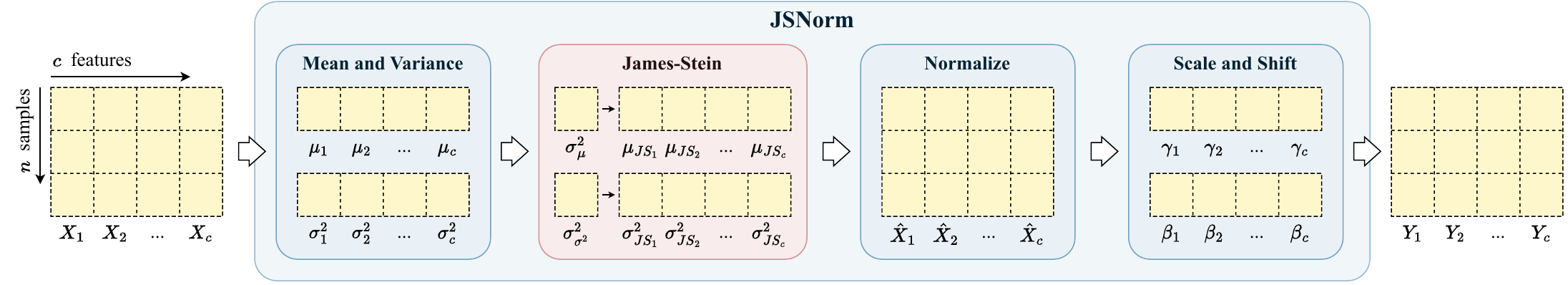}
\caption{The overall structure of our proposed JSNorm for batch normalization with the key components involved at each step.
JSNorm integrates the James-Stein estimator into normalization layers to refine the originally estimated statistics.
This design choice ensures that the computational overhead associated with JSNorm is minimal and practically negligible.}
\label{fig:normalization}
\end{figure*}

\par
\textbf{Admissibility.} Assume $x$ is distributed according to $p(x|\theta)$, with $\theta$ being a member of the set $\Theta$.
Consider $\hat{\theta}$ as an estimator for $\theta$ and let $R(\hat{\theta}, \theta)$ represent the risk of using $\hat{\theta}$, which is calculated based on a specific loss function.
The risk is defined by $R(\hat{\theta}, \theta) = \mathbb{E}[\ell(\hat{\theta}, \theta)]$, where $\ell$ signifies the loss function and the expectation is taken over $x$ drawn from $p(x|\theta)$, with $\hat{\theta}$ being a derived function of $x$.
An estimator is deemed 'inadmissible' if there is another estimator, $\Tilde{\theta}$, which performs better; that is, $R(\Tilde{\theta}, \theta) \leq R(\hat{\theta}, \theta)$ for all $\theta$ within $\Theta$, and there's at least one $\theta$ for which the inequality is strict.
If no such dominating estimator exists, then $\hat{\theta}$ is considered 'admissible' \cite{brown1971admissible}.
\par
\textbf{James-Stein.} Calculating the mean of a multivariate normal distribution stands as a key issue in the field of statistics.
Typically, the sample mean is used, which also happens to be the maximum-likelihood estimator.
However, the James-Stein (JS) estimator, which is known to be biased, is used for estimating the mean of $c$ correlated Gaussian distributed random vectors whose means are not known.
The development of this estimator unfolded over two major papers; the initial version was introduced by Charles Stein in 1956 \cite{stein1956inadmissibility}, leading to the surprising discovery that the standard mean estimate is admissible when $c \leq 2$, but becomes inadmissible for $c \geq 3$.
This work suggested an enhancement that involves shrinking the sample means towards a central vector of means, a concept often cited as Stein's example or paradox \cite{wikipedia_james}.
\par
\textbf{Normalization Layers.} 
The primary objective of normalization techniques is to enhance training stability and facilitate the design of network architectures.
While various normalization layers are available \cite{ulyanov2016instance,wu2018group,salimans2016weight,miyato2018spectral,huang2020normalization}, this paper specifically focuses on batch normalization and layer normalization.
These two normalization techniques are widely utilized in computer vision networks and serve as the most common choices within the field.
\par
Batch Normalization (BatchNorm or BN) \cite{ioffe2015batch} is a key deep learning technique that enhances computer vision applications by normalizing feature-maps using mean and variance computed over batches.
This aids optimization \cite{santurkar2018does}, promotes convergence in deep networks \cite{bjorck2018understanding}, and reduces the number of iterations to converge, improving performance \cite{khoshsirat2023empowering}.
\par
Layer Normalization (LayerNorm or LN) \cite{ba2016layer} is a different approach that addresses training issues in Recurrent Neural Networks \cite{ba2016layer}.
It normalizes statistics within a single sample, making it independent of batch size, and was specifically designed to work with the variable statistics of recurrent neurons.
It is a critical component in transformer networks \cite{vaswani2017attention,liu2021swin}.
\par
\textbf{Contributions.} Our contributions in this study are three-fold:
\begin{itemize}
\item \textbf{Identification of Inadmissible Estimators:}
We establish that widely-used normalization layers in deep learning networks employ inadmissible estimators for calculating mean and variance.
This insight draws attention to potential limitations in the current practice.
\item \textbf{Introduction of Improved Normalization Methods:}
We present an innovative approach wherein the James-Stein estimator is adapted to more accurately estimate the mean and variance in normalization layers.
This is a significant contribution, as it bridges the gap between classical statistical methods and contemporary deep learning techniques.
\item \textbf{Empirical Validation Across Domains:}
Through extensive experimentation, we substantiate the efficacy of our proposed method across three distinct computer vision tasks.
Additionally, we conduct a series of analyses to ascertain the robustness and performance enhancements attributed to our approach under varying configurations.
\end{itemize}

\begin{table*}
\centering
\setlength\tabcolsep{10pt}
\renewcommand{\arraystretch}{2.0}
\begin{tabular}{lll}
 \hline
 Batch Normalization & Layer Normalization & Description \\
 \hline
 $\mu_\mathcal{B} = \frac{1}{n \times h \times w} \sum_{n,h,w} x_{i,j,k}$ &
 $\mu_\mathcal{B} = \frac{1}{h \times w} \sum_{h,w} x_{i,j}$ &
 Calculating the corresponding mean \\
 $\sigma_\mathcal{B}^2 = \frac{1}{n \times h \times w} \sum_{n,h,w} (x_{i,j,k} - \mu_\mathcal{B})^2$ &
 $\sigma_\mathcal{B}^2 = \frac{1}{h \times w} \sum_{h,w} (x_{i,j} - \mu_\mathcal{B})^2$ &
 Calculating the corresponding variance \\
 $\mu_{\mu_\mathcal{B}} = \frac{1}{c} \sum_{c} {\mu_\mathcal{B}}_i$ &
 $\mu_{\mu_\mathcal{B}} = \frac{1}{c} \sum_{c} {\mu_\mathcal{B}}_i$ &
 Mean of the estimated means \\
 $\sigma_{\mu_\mathcal{B}}^2 = \frac{1}{c} \sum_{c} ({\mu_\mathcal{B}}_i - \mu_{\mu_\mathcal{B}})^2$ &
 $\sigma_{\mu_\mathcal{B}}^2 = \frac{1}{c} \sum_{c} ({\mu_\mathcal{B}}_i - \mu_{\mu_\mathcal{B}})^2$ &
 Variance of the estimated means \\
 $\mu_\mathcal{JS} = \left(1 - \frac{(c - 2) \sigma_{\mu_\mathcal{B}}^2}{\|\mu_\mathcal{B}\|^2_2}\right) \mu_\mathcal{B}$ &
 $\mu_\mathcal{JS} = \left(1 - \frac{(c - 2) \sigma_{\mu_\mathcal{B}}^2}{\|\mu_\mathcal{B}\|^2_2}\right) \mu_\mathcal{B}$ &
 JS estimation of the mean \\
 $\mu_{\sigma_\mathcal{B}^2} = \frac{1}{c} \sum_{c} {\sigma_\mathcal{B}^2}_i$ &
 $\mu_{\sigma_\mathcal{B}^2} = \frac{1}{c} \sum_{c} {\sigma_\mathcal{B}^2}_i$ &
 Mean of the estimated variances \\
 $\sigma_{\sigma_\mathcal{B}^2}^2 = \frac{1}{c} \sum_{c} ({\sigma_\mathcal{B}^2}_i - \mu_{\sigma_\mathcal{B}^2})^2$ &
 $\sigma_{\sigma_\mathcal{B}^2}^2 = \frac{1}{c} \sum_{c} ({\sigma_\mathcal{B}^2}_i - \mu_{\sigma_\mathcal{B}^2})^2$ &
 Variance of the estimated variances \\
 $\sigma_\mathcal{JS}^2 = \left(1 - \frac{(c - 2) \sigma_{\sigma_\mathcal{B}^2}^2}{\|\sigma_\mathcal{B}^2\|^2_2}\right) \sigma_\mathcal{B}^2$ &
 $\sigma_\mathcal{JS}^2 = \left(1 - \frac{(c - 2) \sigma_{\sigma_\mathcal{B}^2}^2}{\|\sigma_\mathcal{B}^2\|^2_2}\right) \sigma_\mathcal{B}^2$ &
 JS estimation of the variance \\
 $\hat{x} = \frac{x - \mu_\mathcal{JS}}{\sqrt{\sigma_\mathcal{JS}^2 + \epsilon}}$ &
 $\hat{x} = \frac{x - \mu_\mathcal{JS}}{\sqrt{\sigma_\mathcal{JS}^2 + \epsilon}}$ &
 Standardization using the JS estimations \\
 $y = \gamma \hat{x} + \beta$ &
 $y = \gamma \hat{x} + \beta$ &
 Scale and shift \\
 \hline
\end{tabular}
\caption{Our proposed batch normalization and layer normalization which integrate the James-Stein estimator.
Given an input feature-map $x \in \mathbb{R}^{n \times c \times h \times w}$, for batch normalization, each channel ($c$) is processed separately, and for layer normalization, each sample ($n$).
The operations are done in order, from top to bottom.
}
\label{tab:james-stein}
\end{table*}

\section{Related Work}
The James-Stein estimator has motivated a rich literature on the theme of ``shrinkage'' in statistics.
Just a small selection of examples include LASSO \cite{tibshirani1996regression}, ridge regression \cite{hoerl1970ridge}, the Ledoit-Wolf covariance estimator \cite{ledoit2004well} and Elastic Net \cite{zou2005regularization}.
Excellent textbook treatments of the concepts behind Stein's paradox and James-Stein shrinkage include \cite{gruber2017improving,fourdrinier2018shrinkage}.
\par
Before batch normalization became prevalent, various types of normalization layers were already being utilized in deep neural networks.
Local Response Normalization (LRN) featured in AlexNet \cite{krizhevsky2017imagenet} and was adopted by subsequent models \cite{zeiler2014visualizing,sermanet2013overfeat,khoshsirat2023transformer,hosseini2022application}.
LRN normalizes the values around a given pixel within a specified neighborhood.
\par
Batch Normalization (BN) \cite{ioffe2015batch}, introduced later, applies a more comprehensive form of normalization across the entire batch of data and suggests applying this technique across all layers of the network.
In the wake of BN, other normalization approaches \cite{ba2016layer,ulyanov2016instance,khoshsirat2023sentence,khoshsiratembedding} were developed that do not rely on the batch dimension.
For example, Layer Normalization (LN) \cite{ba2016layer} adjusts the data across the channel dimension, while Instance Normalization (IN) \cite{ulyanov2016instance} carries out a batch normalization-like process for each individual instance.
In a different approach from normalizing data, Weight Normalization \cite{salimans2016weight} focuses on normalizing the weights of the filters within the network.
Group Normalization (GN) \cite{wu2018group} generalizes LN, dividing the neurons into groups and standardizing the layer input within the neurons of each group for each sample independently.
\par
A diverse set of works aiming at improving normalization layers exist.
Decorrelated Batch Normalization \cite{huang2018decorrelated} employs whitening techniques to solve the so-called ``stochastic axis swapping'' problem.
In \cite{xu2019understanding}, a simple version of the layer normalization is introduced that removes the bias and gain parameters to cope with over-fitting.
Batch group normalization \cite{zhou2020batch} extends the grouping mechanism of GN from being over only channels dimension to being over both channels and batch dimensions.
MoBN \cite{salimans2016weight} is a mean-only batch normalization that performs centering only along the batch dimension.
It performs well when combined with weight normalization.
A scaling-only batch normalization is proposed in \cite{yan2020towards} that performs well in training with a small batch size.
\par
Although the James-Stein estimator has not directly been used in deep learning, some methods use this estimator alongside deep learning.
In \cite{angjelichinoski2021deep}, the James-Stein estimator is used for feature extraction before feeding data to a deep neural network.
This method combines Pinsker's theorem with James-Stein to leverage the advantages of non-parametric regression to use deep learning in limited data setups.
C-SURE \cite{chakraborty2020c} is a novel shrinkage estimator based on the James-Stein estimator for the complex-valued manifold.
It is incorporated in a complex-valued classifier network for estimating the statistical mean of the per-class wFM features.
\par
Our research stands at the forefront by being the first study to investigate the admissibility of estimators within normalization layers in deep learning.
Furthermore, we break new ground by introducing an innovative methodology to address this issue, thereby pioneering a potentially transformative approach in the field.

\section{Method}
In this section, we begin by providing a concise overview of batch normalization \cite{ioffe2015batch} and layer normalization \cite{ba2016layer} techniques.
We then delve into our underlying motivation for proposing modified versions of these normalization methods.
Finally, we present our novel approaches to normalization.
\par
\textbf{Batch Normalization.}
The batch normalization layer performs feature standardization within each batch by normalizing each feature individually, followed by learning a shared weight and bias for the entire batch.
Formally, given the input to a batch normalization layer $\mathbf{X} \in \mathbb{R}^{n \times c}$, where $n$ denotes the batch size, and $c$ is the feature dimension, batch normalization layer first normalizes each feature $i \in \{1 \dots c\}$:
\begin{equation} \label{eq:batch-normalization}
\hat{x}_i = \frac{x_i - \mathbb{E}[x_i]}{\sqrt{Var[x_i]}},
\end{equation}
where the expectation and variance are computed over the training data set.
Next, for each feature dimension, a pair of parameters $\gamma_i$ and $\beta_i$ are used to scale and shift the normalized value:
\begin{equation}
y_i = \gamma_i \hat{x}_i + \beta_i.
\end{equation}
These parameters are learned during the training.

\begin{table}[t]
\centering
\setlength\tabcolsep{5pt}
\begin{tabular}{l|cc}
\hline
Method & Original & JSNorm \\
\hline
ResNet-18 \cite{he2016deep} & 69.7 & \textbf{71.0}\textpm 0.3 \textcolor{Green}{(+1.3)} \\
ResNet-50 \cite{he2016deep} & 76.1 & \textbf{77.2}\textpm 0.3 \textcolor{Green}{(+1.1)} \\
ResNet-152 \cite{he2016deep} & 77.9 & \textbf{78.9}\textpm 0.2 \textcolor{Green}{(+1.0)} \\
EfficientNet-B1 \cite{tan2019efficientnet} & 79.2 & \textbf{80.2}\textpm 0.2 \textcolor{Green}{(+1.0)} \\
EfficientNet-B3 \cite{tan2019efficientnet} & 81.7 & \textbf{82.6}\textpm 0.1 \textcolor{Green}{(+0.9)} \\
EfficientNet-B5 \cite{tan2019efficientnet} & 83.7 & \textbf{84.6}\textpm 0.1 \textcolor{Green}{(+0.9)} \\
GENet-light \cite{lin2020neural} & 75.7 & \textbf{77.2}\textpm 0.3 \textcolor{Green}{(+1.5)} \\
\hline
SwinV2-T, win8x8 \cite{liu2022swin} & 81.8 & \textbf{82.8}\textpm 0.2 \textcolor{Green}{(+1.0)} \\
SwinV2-T, win16x16 \cite{liu2022swin} & 82.8 & \textbf{83.7}\textpm 0.2 \textcolor{Green}{(+0.9)} \\
SwinV2-S, win8x8 \cite{liu2022swin} & 83.7 & \textbf{84.6}\textpm 0.1 \textcolor{Green}{(+0.9)} \\
\hline
\end{tabular}
\caption{Comparison of top-1 accuracy on ImageNet with mean\textpm std in five random evaluations.
Notably, all the networks were trained without any fine-tuning of hyper-parameters.}
\label{tab:image-classification}
\end{table}

\par
\textbf{Layer Normalization.}
Similar to batch normalization, layer normalization begins by standardizing the input.
However, unlike batch normalization, layer normalization operates independently on each sample, without relying on the batch size.
As a result, the standardization process occurs individually for each sample.

Therefore, for each sample $i \in \{1 \dots n\}$:
\begin{equation} \label{eq:layer-normalization}
\hat{x}_i = \frac{x_i - \mu_{x_i}}{\sqrt{Var[x_i]}}.
\end{equation}
Similar to batch normalization, it is followed by learning a pair of parameters $\gamma_i$ and $\beta_i$:
\begin{equation}
y_i = \gamma_i \hat{x}_i + \beta_i.
\end{equation}
\par
\textbf{Motivation.}
One notable concern pertaining to Equations \ref{eq:batch-normalization} and \ref{eq:layer-normalization} lies in the estimation of the mean and variance.
The conventional approach suggests independently calculating the mean and variance using ``usual estimators''.
For batch normalization:
\begin{equation}
\mathbb{E}[x_i] = \frac{1}{n} \sum_{j=1}^{n} x_{i,j},
\end{equation}
\begin{equation}
Var[x_i] = \frac{1}{n} \sum_{j=1}^{n} (x_{i,j} - \mathbb{E}[x_i])^2,
\end{equation}
and for layer normalization:
\begin{equation}
\mu_{x_i} = \frac{1}{c} \sum_{j=1}^{c} x_{i,j},
\end{equation}
\begin{equation}
Var[x_i] = \frac{1}{c} \sum_{j=1}^{c} (x_{i,j} - \mu_{x_i})^2,
\end{equation}
are the estimators.
\par
Given that all the features contribute to a shared loss function, according to Stein's paradox \cite{stein1956inadmissibility}, these estimators are inadmissible when $c \geq 3$.
Notably, in computer vision networks, it is consistently observed that $c \geq 3$.
To address this, we propose a novel method to adopt admissible shrinkage estimators, which effectively enhance the estimation of the mean and variance in both normalization layers.

\begin{table}[t]
\centering
\setlength\tabcolsep{7pt}
\begin{tabular}{l|cc}
 \hline
  Method & Original & JSNorm \\
 \hline
 HRNetV2 \cite{wang2019deep} & 81.6 & \textbf{83.0}\textpm 0.4 \textcolor{Green}{(+1.4)} \\
 HRNetV2+OCR \cite{yuan2020object} & 83.0 & \textbf{84.2}\textpm 0.2 \textcolor{Green}{(+1.2)} \\
 HRNetV2+OCR\textsuperscript{\dag} \cite{yuan2020object} & 84.2 & \textbf{85.3}\textpm 0.1 \textcolor{Green}{(+1.1)} \\
 EfficientPS\textsuperscript{\dag} \cite{mohan2021efficientps} & 84.2 & \textbf{85.3}\textpm 0.2 \textcolor{Green}{(+1.1)} \\
 Lawin \cite{yan2022lawin} & 84.4 & \textbf{85.4}\textpm 0.1 \textcolor{Green}{(+1.0)} \\
 \hline
\end{tabular}
\caption{Evaluation of mIoU scores on the Cityscapes test set.
We use \dag ~to mark methods pre-trained on Mapillary Vistas dataset \cite{neuhold2017mapillary}.}
\label{tab:semantic-segmentation}
\end{table}

\begin{table*}[t]
\centering
\setlength\tabcolsep{15pt}
\begin{tabular}{l|cc|cc}
 \hline
 & \multicolumn{2}{c}{ScanObjectNN} & \multicolumn{2}{|c}{ModelNet40} \\
  Method & Original & JSNorm & Original & JSNorm \\
 \hline
 PointNet \cite{qi2017pointnet} & 68.2 & \textbf{70.0}\textpm 0.4 \textcolor{Green}{(+1.8)} & 89.2 & \textbf{90.8}\textpm 0.3 \textcolor{Green}{(+1.6)} \\
 PointNet++ \cite{qi2017pointnet++} & 77.9 & \textbf{79.5}\textpm 0.3 \textcolor{Green}{(+1.6)} & 91.9 & \textbf{93.5}\textpm 0.4 \textcolor{Green}{(+1.6)} \\
 DGCNN \cite{wang2019dynamic} & 81.9 & \textbf{83.3}\textpm 0.1 \textcolor{Green}{(+1.4)} & 93.5 & \textbf{94.8}\textpm 0.2 \textcolor{Green}{(+1.3)} \\
 Point-BERT \cite{yu2022point} & 83.1 & \textbf{84.3}\textpm 0.2 \textcolor{Green}{(+1.2)} & 93.8 & \textbf{94.8}\textpm 0.1 \textcolor{Green}{(+1.0)} \\
 PointNeXt-S \cite{qian2022pointnext} & 87.7 & \textbf{88.8}\textpm 0.1 \textcolor{Green}{(+1.1)} & 93.2 & \textbf{94.2}\textpm 0.1 \textcolor{Green}{(+1.0)} \\
 \hline
\end{tabular}
\caption{Evaluation of classification accuracy across two 3D datasets: ScanObjectNN and ModelNet40.}
\label{tab:3d-classification}
\end{table*}

\par
\textbf{James-Stein.}
Let $\mathbf{X} = \{\mathbf{x}_1, \mathbf{x}_2, ..., \mathbf{x}_c\}$ with unknown means $\boldsymbol{\theta} = \{\theta_1, \theta_2, ..., \theta_c\}$ and estimates $\boldsymbol{\hat{\theta}} = \{\hat{\theta}_1, \hat{\theta}_2, ..., \hat{\theta}_c\}$.
The basic formula for the James-Stein estimator is:
\begin{equation}
\boldsymbol{\hat{\theta}}_{JS} = \boldsymbol{\hat{\theta}} + s(\mu_{\boldsymbol{\hat{\theta}}} - \boldsymbol{\hat{\theta}}),
\end{equation}
where $\mu_{\boldsymbol{\hat{\theta}}} - \boldsymbol{\hat{\theta}}$ is the difference between the total mean (average of averages) and each individual estimated mean, and $s$ is a shrinking factor.
Among the numerous perspectives that motivate the James-Stein estimator, the empirical Bayes perspective \cite{efron1975data} is exquisite.
Taking a Gaussian prior on the unknown means leads us to the following formula \cite{james1961estimation,james1992estimation}:
\begin{equation} \label{eq:james-stein-full}
\boldsymbol{\hat{\theta}}_{JS} = \left(1 - \frac{(c - 2) \sigma^2}{\|\boldsymbol{\hat{\theta}} - \mathbf{v}\|^2_2}\right) (\boldsymbol{\hat{\theta}} - \mathbf{v}) + \mathbf{v},
\end{equation}
where $\| \cdot \|_2$ denotes the $L_2$ norm of the argument, $\sigma^2$ is the variance, $\mathbf{v}$ is an arbitrary fixed vector that shows the shrinkage direction, and $c \geq 3$.
Setting $\mathbf{v} = \mathbf{0}$ results the following:
\begin{equation} \label{eq:james-stein}
\boldsymbol{\hat{\theta}}_{JS} = \left(1 - \frac{(c - 2) \sigma^2}{\|\boldsymbol{\hat{\theta}}\|^2_2}\right) \boldsymbol{\hat{\theta}}.
\end{equation}
The above estimator shrinks the estimates towards the origin $\mathbf{0}$.
\par
Commonly in transformer networks for computer vision \cite{vaswani2017attention,liu2021swin,liu2022swin}, the statistics for layer normalization are calculated on the patch size dimensions $h$, and $w$.
We take advantage of this design and use the James-Stein estimator such that our layer normalization stays independent of batch size, and each sample is processed independently.
\par
In this paper, we employ Equation \ref{eq:james-stein} to avoid the `mean shift' problem \cite{brock2021characterizing,brock2021high}.
To integrate this equation into normalization layers, we substitute the $\boldsymbol{\hat{\theta}}$ in Equation \ref{eq:james-stein} with the estimated mean and variance obtained through the original method.
By applying the James-Stein estimator to the estimated statistics from the original method, the additional processing required is minimal and can be considered negligible.
Consequently, we utilize the James-Stein estimator for both the mean and variance in the normalization layers.
For batch normalization, $\mathbb{E}[x]$ and $Var[x]$ are vectors of length $c$ (for the whole batch).
Therefore, they can directly be used in place of $\boldsymbol{\hat{\theta}}$.
For layer normalization, $\mu_x$ and $Var[x]$ are in the form of $\mathbb{R}^{n \times c}$, and since in layer normalization each sample should be processed independently, each vector from the second dimension $c$ is separately used in place of $\boldsymbol{\hat{\theta}}$.
Table \ref{tab:james-stein} shows the detailed definition of our proposed normalization layers, while Figure \ref{fig:normalization} depicts the overview of JSNorm for batch normalization.
\par
\subsection{Gaussian Prior}
Employing a variant of the James-Stein estimator that assumes a Gaussian prior might raise queries regarding the general applicability of our methodology.
This concern can be addressed through two key observations:
\begin{itemize}
\item \textbf{Alignment with Normal Distribution in Normalization Layers:}
Normalization layers aim to standardize the features-maps within a layer and make them resemble a standard Gaussian distribution with a mean of zero and a standard deviation of one.
Thus, adopting a Gaussian prior aligns with the intrinsic characteristics of the feature-maps and does not substantially deviate their distribution within the network.
Moreover, numerous approaches to developing normalization-free networks \cite{brock2021characterizing,brock2021high} incorporate Gaussian weight initialization and standardization to promote a Gaussian-like distribution for feature-maps.
This underscores the practical advantages of adopting a Gaussian prior.
\item \textbf{Distinction Between Feature-Maps and Input Distributions:}
It is crucial to acknowledge that the distribution of the feature-maps within the neural network need not mirror that of the input data.
The succession of transformations applied by the network layers often results in an evolution of the input data distribution.
As the network trains, it learns to change data representation in a manner useful for the specific task.
Therefore, the efficacy of our method is not strictly tethered to the distribution of inputs.
\end{itemize}

\section{Experiments}
We conduct a comprehensive evaluation of our proposed method across various computer vision tasks.
For each task, we utilize well-established state-of-the-art networks with readily available implementations.
The sole modification we introduce involves replacing the normalization layers with our proposed batch normalization or layer normalization counterparts.
As a result, all other hyper-parameters and training configurations remain consistent with those outlined in the original papers.

\subsection{Image Classification}
Our evaluation of the proposed method for image classification involves using the ImageNet dataset \cite{russakovsky2015imagenet}, which consists of 1.28 million training images and 50,000 validation images across 1,000 classes.
We train the networks from scratch and report the top-1 accuracy achieved.
The results are presented and compared in Table \ref{tab:image-classification}, where we assess the performance of our modified batch normalization on various model sizes from the ResNet \cite{he2016deep}, EfficientNet \cite{tan2019efficientnet}, and GENet \cite{lin2020neural} families, as well as the SwinV2 \cite{liu2022swin} model with layer normalization.
\par
Across the ResNet models, we observe a maximum accuracy improvement of 1.3\% for ResNet-18, while the larger models show slightly lesser improvement.
This trend holds true for other network architectures as well.
For GENet-light, we achieve a maximum accuracy improvement of 1.5\%, and EfficientNet-B5 demonstrates a minimum accuracy improvement of 0.9\%.
While it is true that the gains are more pronounced for ResNet-18 or GENet-light, it is important to note that our method also yields steady improvements on larger networks, demonstrating its broad applicability.
It is generally accepted that larger networks, which already attain higher accuracy levels and reach performance saturation, exhibit less incremental improvement when more regularizers or data augmentation techniques are introduced.
This behavior is consistent with that of the JSNorm, which functions as a regularizer and is inherently similar to other regularizers.
The smaller improvements observed in larger networks can be attributed to the presence of existing regularizers, which can limit the additional boost provided by introducing another regularizer.
This is one of the motivations for performing the Regularization Effect ablation study (Section \ref{sec:regularization-effect}).
\par
The findings in this section highlight the compatibility of our proposed JSNorm with both convolutional and transformer networks, indicating its effectiveness in improving model performance.

\subsection{Semantic Segmentation}
The Cityscapes dataset \cite{cordts2016cityscapes} serves as our primary evaluation dataset for semantic segmentation, consisting of 5,000 high-quality street images with pixel-level annotations.
These finely annotated images are divided into subsets of 2,975 for training, 500 for validation, and 1,525 for testing.
Additionally, the dataset includes an additional 20,000 coarsely annotated images.
It contains 30 classes, with 19 classes used for performance assessment.
\par
Our experiments involve the use of HRNetV2 \cite{wang2019deep} and its augmented version HRNetV2+OCR \cite{yuan2020object}, as well as EfficientPS \cite{mohan2021efficientps} and Lawin \cite{yan2022lawin}.
HRNetV2 is a fully convolutional network that maintains high-resolution representations throughout the network.
EfficientPS employs EfficientNet as a backbone for semantic and panoptic segmentation.
Meanwhile, Lawin is a multi-scale transformer network that utilizes a window attention mechanism.
In our evaluation, we replace the normalization layers in the aforementioned networks with our proposed normalization layer and compare the results.
\par
Table \ref{tab:semantic-segmentation} presents the mean intersection over union (mIoU) measure for these models on the Cityscapes dataset.
Our improved batch normalization demonstrates notable enhancements for HRNetV2, achieving an improvement of 1.4\%.
Additionally, we observe a minimum improvement of 1.0\%, which boosts Lawin's accuracy to reach 85.4\% on this dataset.
These results showcase the efficacy of our proposed normalization approach in improving the segmentation performance across different models.

\begin{table}[t]
\centering
\setlength\tabcolsep{4pt}
\begin{tabular}{l|ccc}
\hline
Method & Original & Ridge & LASSO \\
\hline
ResNet-18 \cite{he2016deep} & 69.7 & 70.1 \textcolor{Green}{(+0.4)} & 70.1 \textcolor{Green}{(+0.4)} \\
ResNet-152 \cite{he2016deep} & 77.9 & 78.0 \textcolor{Green}{(+0.1)} & 78.1 \textcolor{Green}{(+0.2)} \\
EfficientNet-B1 \cite{tan2019efficientnet} & 79.2 & 79.4 \textcolor{Green}{(+0.2)} & 79.5 \textcolor{Green}{(+0.3)} \\
EfficientNet-B5 \cite{tan2019efficientnet} & 83.7 & 83.7 (+0.0) & 83.8 \textcolor{Green}{(+0.1)} \\
GENet-light \cite{lin2020neural} & 75.7 & 76.0 \textcolor{Green}{(+0.3)} & 76.0 \textcolor{Green}{(+0.3)} \\
\hline
SwinV2-T \cite{liu2022swin} & 81.8 & 81.9 \textcolor{Green}{(+0.1)} & 81.9 \textcolor{Green}{(+0.1)} \\
SwinV2-S \cite{liu2022swin} & 83.7 & 83.7 (+0.0) & 83.7 (+0.0) \\
\hline
\end{tabular}
\caption{Comparative analysis of two widely used shrinkage estimators: Ridge and LASSO.
The figures represent the Top-1 accuracy on the ImageNet dataset.}
\label{tab:shrinkage-estimators}
\end{table}

\subsection{3D Object Classification}
A 3D point cloud, which comprises an unordered collection of 3D points, necessitates distinct network architectures compared to those tailored for 2D images.
This, in turn, creates an alternative platform for evaluation.
We have conducted experiments utilizing two datasets tailored for 3D object classification.
The first dataset, ScanObjectNN \cite{uy2019revisiting}, is derived from real 3D scenes, and its inherent complexity, augmented by occlusions and noise, poses substantial challenges for prevailing 3D classification methodologies.
ScanObjectNN encompasses 2,309 training and 581 testing point clouds, distributed across 15 object classes.
The second dataset, ModelNet40 \cite{wu20153d}, is widely recognized in the realm of 3D object classification and comprises synthetic object point clouds.
The dataset contains 12,311 CAD-generated meshes categorized into 40 classes, and is partitioned into 9,843 training and 2,468 testing samples.
We assessed the efficacy of our proposed technique on five models, four incorporating batch normalization - PointNet \cite{qi2017pointnet}, PointNet++ \cite{qi2017pointnet++}, DGCNN \cite{wang2019dynamic}, PointNeXt \cite{qian2022pointnext} - and one employing layer normalization, namely Point-BERT \cite{yu2022point}.
\par
Table \ref{tab:3d-classification} presents the classification accuracy for both ScanObjectNN and ModelNet40 datasets.
Remarkably, by employing our enhanced JSNorm, PointNet attains a substantial improvement in classification accuracy, increasing by 1.8\% on the ScanObjectNN dataset.
Even at the lower end of the spectrum, Point-BERT on ModelNet40 exhibits a noteworthy enhancement in accuracy by 1.0\%.
These findings underscore the versatility of our proposed approach, indicating that its application extends beyond 2D image networks and is adaptable to a diverse array of data formats and network architectures.

\section{Extra Studies}
In addition to our primary study, we conduct an additional study that compares two alternative shrinkage estimators.
Furthermore, we carry out three ablation studies to gain insights into the impact of regularization, shrinkage, and batch size on the performance of our proposed method.

\subsection{Ridge and LASSO Estimators}
In this study, we assess two widely utilized shrinkage estimators, namely Ridge \cite{hoerl1970ridge} and LASSO \cite{tibshirani1996regression}, to determine whether alternative shrinkage estimators possess the capacity to enhance accuracy.
In terms of the Ridge estimator, $\boldsymbol{\hat{\theta}}$ can be estimated via:
\begin{equation}
\boldsymbol{\hat{\theta}}_{Ridge} = \operatorname*{argmin}_{\boldsymbol{\hat{\theta}}} \left[ \ell(\boldsymbol{\hat{\theta}}, \boldsymbol{\theta}) + \lambda \|\boldsymbol{\hat{\theta}}\|^2_2 \right],
\end{equation}
and in terms of LASSO:
\begin{equation}
\boldsymbol{\hat{\theta}}_{LASSO} = \operatorname*{argmin}_{\boldsymbol{\hat{\theta}}} \left[ \ell(\boldsymbol{\hat{\theta}}, \boldsymbol{\theta}) + \lambda \|\boldsymbol{\hat{\theta}}\|_1 \right],
\end{equation}
where $\lambda$ is called the regularization parameter and controls the amount of shrinkage.
Both the Ridge and LASSO estimators perform regularization of the estimated parameters, with the LASSO estimator offering the added benefit of variable selection, which enhances the interpretability of the model.
Both estimators also introduce shrinkage towards $\mathbf{0}$ as part of their regularization process.

\begin{figure}[t]
\includegraphics[width=\linewidth]{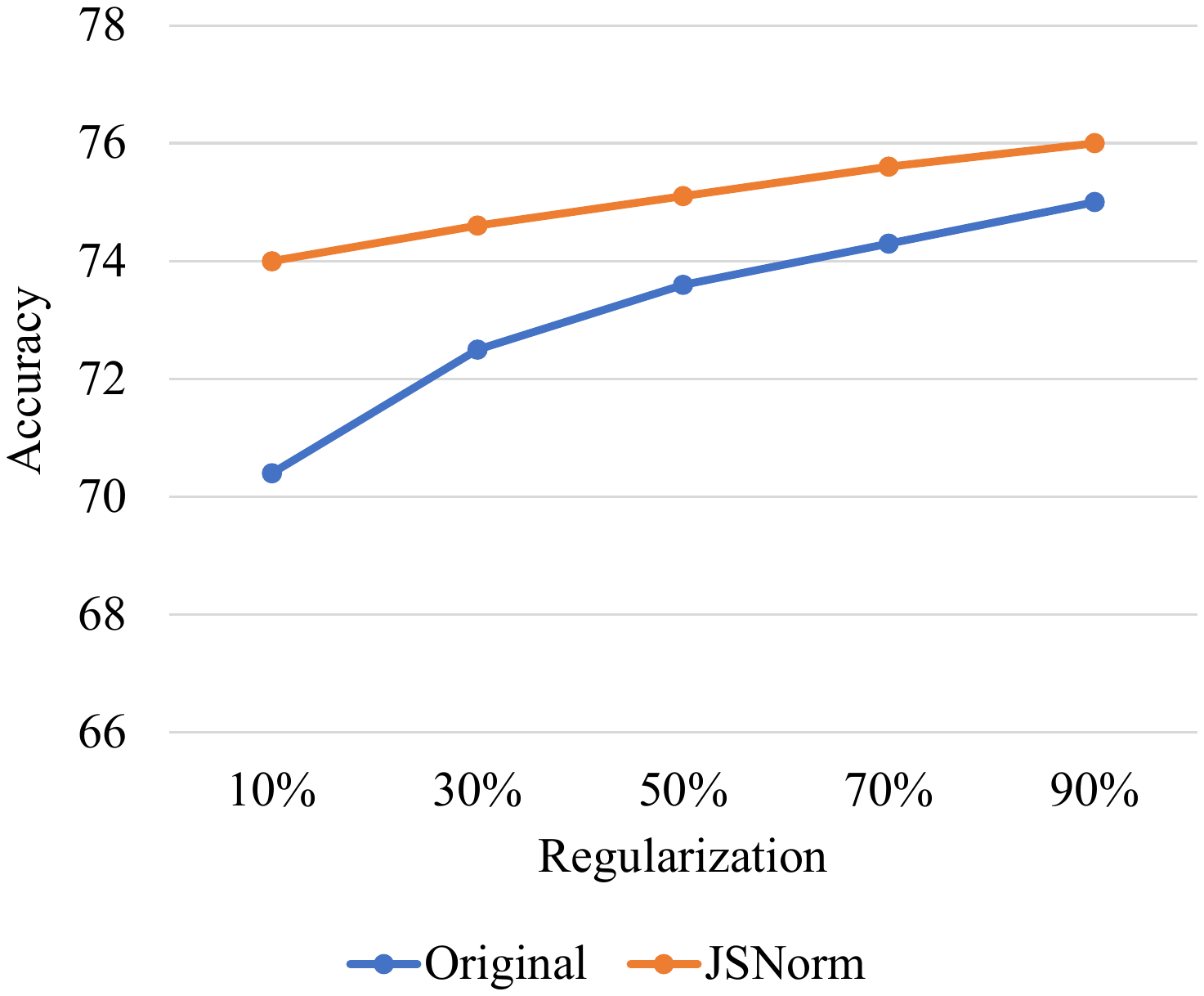}
\caption{Comparative analysis of our enhanced batch normalization performance across various regularization intensities.
JSNorm not only boosts accuracy but also exhibits increased robustness, particularly under lower regularization.}
\label{fig:regularization}
\end{figure}

\begin{figure*}[t]
\includegraphics[width=\linewidth]{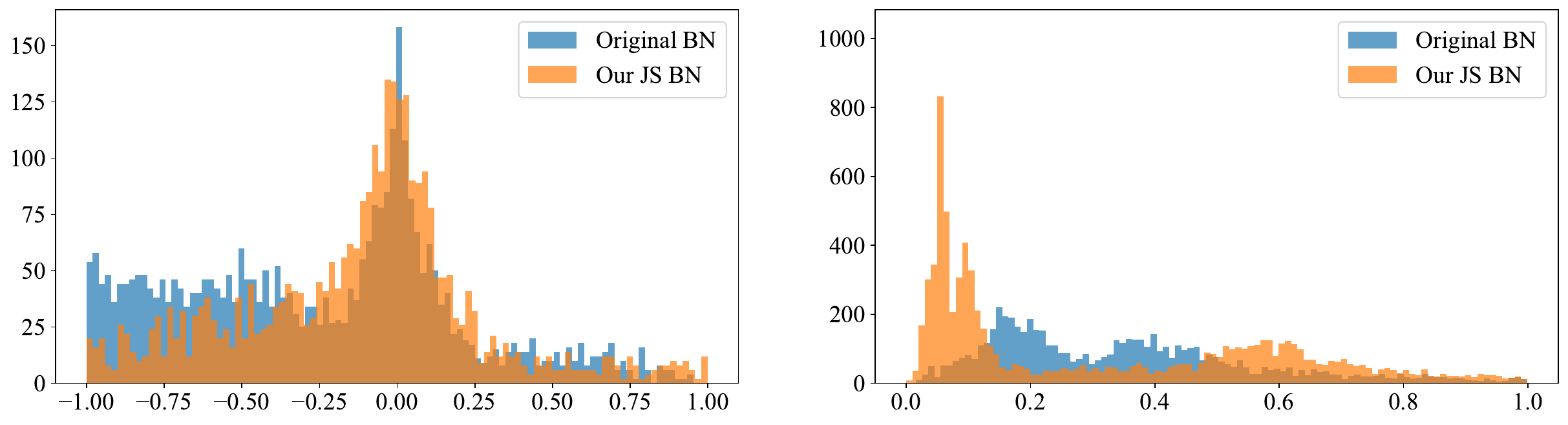}
\caption{The shrinkage effect of JSNorm on ResNet-18 \cite{he2016deep}.
The figures illustrate the distribution of running means (left) and running variances (right) in the original batch normalization compared to JSNorm, showcasing the shrinkage effect induced by our JSNorm layer.}
\label{fig:distribution}
\end{figure*}

\par
We incorporate the Ridge and LASSO estimators as regularization components acting upon the estimated mean and variance within batch normalization, a methodology that can be seamlessly extended to layer normalization as well.
It is imperative to highlight that there is no alteration to the original formulas of batch normalization when employing Ridge and LASSO estimators.
To integrate these regularization components, we modify the training loss function in the following manner:
\begin{equation} \label{eq:loss-function}
\ell_{final} = \ell_{original} + \lambda \sum f(\mu_\mathcal{B}) + f(\sigma_\mathcal{B}^2),
\end{equation}
where $\sum$ is summation over all batch normalization layers and $f$ is the regularization term, $\|\boldsymbol{\cdot}\|^2_2$ for Ridge and $\|\boldsymbol{\cdot}\|_1$ for LASSO.
Since the value of $\ell_{original}$ can be very much larger or smaller than $\sum f(\mu_\mathcal{B}) + f(\sigma_\mathcal{B}^2)$ for different tasks and networks, $\lambda$ needs to be tuned accordingly.
To solve this problem, we re-scale $\lambda$ proportionally to the value of the regularization part and $\ell_{original}$:
\begin{equation}
\lambda = \lambda_{original} (\frac{\ell_{original}}{\sum f(\mu_\mathcal{B}) + f(\sigma_\mathcal{B}^2)}).
\end{equation}
This way, $\lambda_{original}$ is the hyper-parameter that should be chosen.
Re-scaling $\lambda$ happens outside the computational graph to prevent it from affecting gradient calculation.
\par
We subject the regularized normalization layers to evaluation within the context of the image classification task.
Table \ref{tab:shrinkage-estimators} presents the top-1 accuracy on the ImageNet dataset for several distinct networks.
While Ridge and LASSO contribute to enhanced performance, they do not match the level of improvement achieved by the James-Stein estimator.
Given that the James-Stein estimator is predicated on the assumption of a Gaussian distribution underlying the data, this may account for its superior performance relative to Ridge and LASSO.
In our experimental assessments, LASSO marginally outperforms Ridge in terms of accuracy, though their performances are largely analogous.
The results underscore the capability of various shrinkage estimators to bolster accuracy, albeit not to the same extent as James-Stein.

\begin{table}[t]
\centering
\setlength\tabcolsep{6pt}
\begin{tabular}{c|cccccc}
 \hline
  Method & 32 & 64 & 128 & 256 & 512 & 1024 \\
 \hline
 Original & 75.1 & 75.4 & 75.6 & 75.7 & 75.6 & 75.5 \\
 JSNorm & \textbf{76.8} & \textbf{77.0} & \textbf{77.1} & \textbf{77.1} & \textbf{77.0} & \textbf{76.9} \\
 $\Delta$ & \textcolor{Green}{+1.7} & \textcolor{Green}{+1.6} & \textcolor{Green}{+1.5} & \textcolor{Green}{+1.4} & \textcolor{Green}{+1.4} & \textcolor{Green}{+1.4} \\
 \hline
\end{tabular}
\caption{Impact of varying training batch sizes on ImageNet accuracy.
Our JS batch normalization improves accuracy and demonstrates superior robustness across different batch sizes.}
\label{tab:batch-size}
\end{table}

\subsection{Regularization Effect} \label{sec:regularization-effect}
In the scholarly domain, it is well-established that shrinkage estimators inherently exhibit regularization effects \cite{hoerl1970ridge,tibshirani1996regression}.
In this investigation, we illuminate the performance characteristics of our novel JSNorm layers across an array of regularization magnitudes.
For this purpose, we deploy three regularization techniques, namely RandAugment \cite{cubuk2020randaugment}, stochastic depth \cite{huang2016deep}, and dropout \cite{srivastava2014dropout}.
We initiate the experiment with a baseline configuration using GENet-light \cite{lin2020neural} and progressively escalate the regularization factors.
Guided by the methodologies delineated in \cite{cubuk2020randaugment} and \cite{tan2021efficientnetv2}, we establish the upper bounds for regularization factors.
The combined regularization is represented in percentage to foster clarity and streamline the visualisation.
We train the network under two scenarios – one employing our enhanced batch normalization and the other without.
\par
Figure \ref{fig:regularization} presents a graphical representation that compares the two approaches across a range of regularization factors.
An intriguing observation is that our method exhibits its most pronounced enhancement in scenarios characterized by low regularization.
Nonetheless, even in scenarios with maximal regularization, our method registers noteworthy accuracy improvements.
This exemplifies the robustness and stability of our proposition across a spectrum of regularization parameters, thereby outclassing the conventional approach.

\subsection{Shrinkage Effect}
Shrinkage estimators exert a shrinkage effect on the estimated parameters.
We are keen to analyze how this shrinkage influences the normalization layers.
To accomplish this, we provide a visual comparison of the distribution of the running statistics within the batch normalization layers of a ResNet-18 model \cite{he2016deep}.
Two networks are trained, one employing the standard batch normalization and the other utilizing our enhanced batch normalization.
\par
Figure \ref{fig:distribution} offers a graphical representation of the histograms of all the running means (left subfigure) and variances (right subfigure), both with and without the incorporation of our JSNorm.
The subfigures illuminate how our JSNorm layer nudges the distributions toward zero.
Pertaining to the running means, the employment of our JSNorm facilitates a distribution that is less skewed and more bell-shaped.
As for the running variances, directing the distribution towards zero does not hinder the network's capability to learn layers with elevated variances.
In essence, it contributes positively to the diversity of the distribution.

\subsection{Batch Size Effect}
Batch size plays a pivotal role in the training process of networks that incorporate batch normalization layers.
Given that our method enhances the estimation of normalization statistics, it is intriguing to investigate its performance across varying batch sizes.
This study is designed to elucidate the behavior of our James-Stein-augmented batch normalization in the context of different batch sizes.
To achieve this, we train a GENet-light \cite{lin2020neural} network on the ImageNet dataset using a range of batch sizes, spanning from 32 to 1024.
We employ the linear learning rate scaling rule \cite{goyal2017accurate} to adjust to the alterations in batch size.
\par
Table \ref{tab:batch-size} presents the findings of this investigation.
Our enhanced batch normalization exhibits optimal performance with smaller batch sizes, attributable to the James-Stein estimator's proficiency in estimating normalization statistics with a limited sample pool.
Notably, even with larger batch sizes, our method surpasses the performance of the original batch normalization by a considerable margin.
Consequently, our JSNorm demonstrates not only an enhancement in accuracy but also exhibits robustness with respect to batch size variations.

\section{Conclusion}
Through the lens of Stein's paradox, we illustrated that normalization layers employ inadmissible estimators, resulting in suboptimal estimations of layer statistics.
To address this issue, we introduced an innovative technique utilizing the James-Stein estimator, which enhances the accuracy of mean and variance estimations.
We evaluated our proposed method rigorously across three distinct computer vision tasks.
Our findings demonstrated that the technique not only bolstered the accuracy of both convolutional and transformer networks but did so without incurring additional computational overhead.
We performed extra studies to unveil that our approach exhibits robustness and is less susceptible to changes in batch size and regularization, leading to consistent improvements in accuracy under diverse configurations.

{\small
\bibliographystyle{ieee_fullname}
\bibliography{references}
}

\title{Supplementary Materials}
\author{}
\maketitle

Our JSNorm layers require the same computational resources as the original normalization layers.
However, the current implementations of normalization layers employ an optimized mathematical expression tailored for computing derivatives during the backpropagation phase.
This optimization is achieved through manual derivation and operates independently of the automatic differentiation mechanisms that are integral to many machine learning frameworks.
\par
In this supplementary material, we present the optimized expression for the derivative of our method.
Specifically, we provide the details for our batch normalization layer, while noting that similar expressions can be used for our layer normalization.
We maintain the same notation as in the main text, with the addition of the symbol $\mathcal{S}$ denoting the sum of the squares of a vector, or $\| \cdot \|^2_2$.
Therefore, $\mathcal{S}_{\mu_\mathcal{B}} = \| \mu_\mathcal{B} \|^2_2$.

\section{Chain Rule Expansions}
Below, we present the partial derivatives derived using the chain rule, starting from the final output $y_i$.
The main three outputs are the partial derivatives of the loss function $\ell$ with respect to $\gamma$, $\beta$, and $x_i$, given an input $x \in \mathbb{R}^{n \times c}$.

\begin{equation}
\frac{\partial \ell}{\partial \gamma} = \frac{\partial \ell}{\partial y_i} \cdot \frac{\partial y_i}{\partial \gamma}
\end{equation}
\begin{equation}
\frac{\partial \ell}{\partial \beta} = \frac{\partial \ell}{\partial y_i} \cdot \frac{\partial y_i}{\partial \beta}
\end{equation}
\begin{equation}
\frac{\partial \ell}{\partial x_i} = 
\frac{\partial \ell}{\partial \hat{x}_i} \cdot \frac{\partial \hat{x}_i}{\partial x_i} + 
\frac{\partial \ell}{\partial \mu_\mathcal{B}} \cdot \frac{\partial \mu_\mathcal{B}}{\partial x_i} + 
\frac{\partial \ell}{\partial \sigma_\mathcal{B}^2} \cdot \frac{\partial \sigma_\mathcal{B}^2}{\partial x_i}
\end{equation}
\begin{equation}
\begin{split}
\frac{\partial \ell}{\partial \mu_\mathcal{B}} & = 
\frac{\partial \ell}{\partial \mu_\mathcal{JS}} \cdot \frac{\partial \mu_\mathcal{JS}}{\partial \mu_\mathcal{B}} +
\frac{\partial \ell}{\partial \mathcal{S}_{\mu_\mathcal{B}}} \cdot \frac{\partial \mathcal{S}_{\mu_\mathcal{B}}}{\partial \mu_\mathcal{B}} +
\frac{\partial \ell}{\partial \sigma_{\mu_\mathcal{B}}^2} \cdot \frac{\partial \sigma_{\mu_\mathcal{B}}^2}{\partial \mu_\mathcal{B}} \\ & +
\frac{\partial \ell}{\partial \mu_{\mu_\mathcal{B}}} \cdot \frac{\partial \mu_{\mu_\mathcal{B}}}{\partial \mu_\mathcal{B}} +
\frac{\partial \ell}{\partial \sigma_\mathcal{B}^2} \cdot \frac{\partial \sigma_\mathcal{B}^2}{\partial \mu_\mathcal{B}}
\end{split}
\end{equation}
\begin{equation}
\begin{split}
\frac{\partial \ell}{\partial \sigma_\mathcal{B}^2} & = 
\frac{\partial \ell}{\partial \sigma_\mathcal{JS}^2} \cdot \frac{\partial \sigma_\mathcal{JS}^2}{\partial \sigma_\mathcal{B}^2} +
\frac{\partial \ell}{\partial \mathcal{S}_{\sigma_\mathcal{B}^2}} \cdot \frac{\partial \mathcal{S}_{\sigma_\mathcal{B}^2}}{\partial \sigma_\mathcal{B}^2} +
\frac{\partial \ell}{\partial \sigma_{\sigma_\mathcal{B}^2}^2} \cdot \frac{\partial \sigma_{\sigma_\mathcal{B}^2}^2}{\partial \sigma_\mathcal{B}^2} \\ & +
\frac{\partial \ell}{\partial \mu_{\sigma_\mathcal{B}^2}} \cdot \frac{\partial \mu_{\sigma_\mathcal{B}^2}}{\partial \sigma_\mathcal{B}^2}
\end{split}
\end{equation}
\begin{equation}
\frac{\partial \ell}{\partial \mathcal{S}_{\mu_\mathcal{B}}} =
\frac{\partial \ell}{\partial \mu_\mathcal{JS}} \cdot \frac{\partial \mu_\mathcal{JS}}{\partial \mathcal{S}_{\mu_\mathcal{B}}}
\end{equation}
\begin{equation}
\frac{\partial \ell}{\partial \mathcal{S}_{\sigma_\mathcal{B}^2}} =
\frac{\partial \ell}{\partial \sigma_\mathcal{JS}^2} \cdot \frac{\partial \sigma_\mathcal{JS}^2}{\partial \mathcal{S}_{\sigma_\mathcal{B}^2}}
\end{equation}
\begin{equation}
\frac{\partial \ell}{\partial \mu_{\mu_\mathcal{B}}} =
\frac{\partial \ell}{\partial \sigma_{\mu_\mathcal{B}}^2} \cdot \frac{\partial \sigma_{\mu_\mathcal{B}}^2}{\partial \mu_{\mu_\mathcal{B}}}
\end{equation}
\begin{equation}
\frac{\partial \ell}{\partial \sigma_{\mu_\mathcal{B}}^2} =
\frac{\partial \ell}{\partial \mu_\mathcal{JS}} \cdot \frac{\partial \mu_\mathcal{JS}}{\partial \sigma_{\mu_\mathcal{B}}^2}
\end{equation}
\begin{equation}
\frac{\partial \ell}{\partial \mu_{\sigma_\mathcal{B}^2}} =
\frac{\partial \ell}{\partial \sigma_{\sigma_\mathcal{B}^2}^2} \cdot \frac{\partial \sigma_{\sigma_\mathcal{B}^2}^2}{\partial \mu_{\sigma_\mathcal{B}^2}}
\end{equation}
\begin{equation}
\frac{\partial \ell}{\partial \sigma_{\sigma_\mathcal{B}^2}^2} =
\frac{\partial \ell}{\partial \sigma_\mathcal{JS}^2} \cdot \frac{\partial \sigma_\mathcal{JS}^2}{\partial \sigma_{\sigma_\mathcal{B}^2}^2}
\end{equation}
\begin{equation}
\frac{\partial \ell}{\partial \mu_\mathcal{JS}} = 
\frac{\partial \ell}{\partial \hat{x}_i} \cdot \frac{\partial \hat{x}_i}{\partial \mu_\mathcal{JS}}
\end{equation}
\begin{equation}
\frac{\partial \ell}{\partial \sigma_\mathcal{JS}^2} = 
\frac{\partial \ell}{\partial \hat{x}_i} \cdot \frac{\partial \hat{x}_i}{\partial \sigma_\mathcal{JS}^2}
\end{equation}

\section{Partial Derivatives}
The actual derivatives for the partials are calculated as follows:
\begin{equation}
\frac{\partial \ell}{\partial \gamma} = \sum\limits_{i=1}^n \frac{\partial \ell}{\partial y_i} \cdot \hat{x}_i
\end{equation}
\begin{equation}
\frac{\partial \ell}{\partial \beta} = \sum\limits_{i=1}^n \frac{\partial \ell}{\partial y_i}
\end{equation}
\begin{equation}
\frac{\partial \ell}{\partial \hat{x}_i} = \frac{\partial \ell}{\partial y_i} \cdot \gamma
\end{equation}
\begin{equation}
\frac{\partial \hat{x}_i}{\partial x_i} = \frac{1}{\sqrt{\sigma_\mathcal{JS}^2 + \epsilon}}
\end{equation}
\begin{equation}
\frac{\partial \hat{x}_i}{\partial \mu_\mathcal{JS}} = \frac{-1}{\sqrt{\sigma_\mathcal{JS}^2 + \epsilon}}
\end{equation}
\begin{equation}
\frac{\partial \hat{x}_i}{\partial \sigma_\mathcal{JS}^2} = -0.5 \sum\limits_{i=1}^n (x_i - \mu_\mathcal{JS}) \cdot (\sigma_\mathcal{JS}^2 + \epsilon)^{-1.5}
\end{equation}
\begin{equation}
\frac{\partial \mu_\mathcal{JS}}{\partial \mu_\mathcal{B}} = 
1 - \frac{(c - 2) \sigma_{\mu_\mathcal{B}}^2}{\mathcal{S}_{\mu_\mathcal{B}}}
\end{equation}
\begin{equation}
\frac{\partial \sigma_\mathcal{JS}^2}{\partial \sigma_\mathcal{B}^2} = 
1 - \frac{(c - 2) \sigma_{\sigma_\mathcal{B}^2}^2}{\mathcal{S}_{\sigma_\mathcal{B}^2}}
\end{equation}
\begin{equation}
\frac{\partial \mu_\mathcal{JS}}{\partial \mathcal{S}_{\mu_\mathcal{B}}} =
\frac{\mu_\mathcal{B} \cdot (c - 2) \cdot \sigma_{\mu_\mathcal{B}}^2}{\mathcal{S}^2_{\mu_\mathcal{B}}}
\end{equation}
\begin{equation}
\frac{\partial \sigma_\mathcal{JS}^2}{\partial \mathcal{S}_{\sigma_\mathcal{B}^2}} =
\frac{\sigma_\mathcal{B}^2 \cdot (c - 2) \cdot \sigma_{\sigma_\mathcal{B}^2}^2}{\mathcal{S}^2_{\sigma_\mathcal{B}^2}}
\end{equation}
\begin{equation}
\frac{\partial \mathcal{S}_{\mu_\mathcal{B}}}{\partial \mu_\mathcal{B}} = 2 \cdot \mu_\mathcal{B}
\end{equation}
\begin{equation}
\frac{\partial \mathcal{S}_{\sigma_\mathcal{B}^2}}{\partial \sigma_\mathcal{B}^2} = 2 \cdot \sigma_\mathcal{B}^2
\end{equation}
\begin{equation}
\frac{\partial \mu_\mathcal{JS}}{\partial \sigma_{\mu_\mathcal{B}}^2} = \frac{- \mu_\mathcal{B} \cdot (c - 2)}{\mathcal{S}_{\mu_\mathcal{B}}}
\end{equation}
\begin{equation}
\frac{\partial \sigma_\mathcal{JS}^2}{\partial \sigma_{\sigma_\mathcal{B}^2}^2} = \frac{- \sigma_\mathcal{B}^2 \cdot (c - 2)}{\mathcal{S}_{\sigma_\mathcal{B}^2}}
\end{equation}
\begin{equation}
\frac{\partial \mu_\mathcal{B}}{\partial x_i} = \frac{1}{n}
\end{equation}
\begin{equation}
\frac{\partial \mu_{\mu_\mathcal{B}}}{\partial \mu_\mathcal{B}} = \frac{1}{c}
\end{equation}
\begin{equation}
\frac{\partial \mu_{\sigma_\mathcal{B}^2}}{\partial \sigma_\mathcal{B}^2} = \frac{1}{c}
\end{equation}
\begin{equation}
\frac{\partial \sigma_\mathcal{B}^2}{\partial x_i} = \frac{2(x_i - \mu_\mathcal{B})}{n}
\end{equation}
\begin{equation}
\frac{\partial \sigma_{\sigma_\mathcal{B}^2}^2}{\partial \sigma_\mathcal{B}^2} = \frac{2(\sigma_\mathcal{B}^2 - \mu_{\sigma_\mathcal{B}^2})}{c}
\end{equation}
\begin{equation}
\frac{\partial \sigma_{\mu_\mathcal{B}}^2}{\partial \mu_\mathcal{B}} = \frac{2(\mu_\mathcal{B} - \mu_{\mu_\mathcal{B}})}{c}
\end{equation}
\begin{equation}
\frac{\partial \sigma_\mathcal{B}^2}{\partial \mu_\mathcal{B}} = \frac{1}{n} \sum\limits_{i=1}^n -2 \cdot (x_i - \mu_\mathcal{B})
\end{equation}
\begin{equation}
\frac{\partial \sigma_{\sigma_\mathcal{B}^2}^2}{\partial \mu_{\sigma_\mathcal{B}^2}} =
\frac{1}{c} \sum\limits_{i=1}^c -2 \cdot ({\sigma_\mathcal{B}^2}_i - \mu_{\sigma_\mathcal{B}^2})
\end{equation}
\begin{equation}
\frac{\partial \sigma_{\mu_\mathcal{B}}^2}{\partial \mu_{\mu_\mathcal{B}}} =
\frac{1}{c} \sum\limits_{i=1}^c -2 \cdot ({\mu_\mathcal{B}}_i - \mu_{\mu_\mathcal{B}})
\end{equation}

\section{Simplifying Expressions}
Multiple expressions can be simplified as follows:
\begin{equation}
\begin{split}
\frac{\partial \sigma_\mathcal{B}^2}{\partial \mu_\mathcal{B}} & =
\frac{1}{n} \sum\limits_{i=1}^n -2 \cdot (x_i - \mu_\mathcal{B})
\\ & = -2 \cdot \left(\frac{1}{n} \sum\limits_{i=1}^n x_i - \frac{1}{n} \sum\limits_{i=1}^n \mu_\mathcal{B} \right)
\\ & = -2 \cdot \left(\mu_\mathcal{B} - \frac{n \cdot \mu_\mathcal{B}}{n} \right)
\\ & = -2 \cdot \left(\mu_\mathcal{B} - \mu_\mathcal{B} \right)
\\ & = 0
\end{split}
\end{equation}
Similarly:
\begin{equation}
\frac{\partial \sigma_{\sigma_\mathcal{B}^2}^2}{\partial \mu_{\sigma_\mathcal{B}^2}} = 0
\end{equation}
\begin{equation}
\frac{\partial \sigma_{\mu_\mathcal{B}}^2}{\partial \mu_{\mu_\mathcal{B}}} = 0
\end{equation}
Therefore:
\begin{equation}
\frac{\partial \ell}{\partial \mu_{\sigma_\mathcal{B}^2}} = 0
\end{equation}
\begin{equation}
\frac{\partial \ell}{\partial \mu_{\mu_\mathcal{B}}} = 0
\end{equation}
By utilizing the above results, the partial derivatives $\frac{\partial \ell}{\partial \mu_\mathcal{B}}$ and $\frac{\partial \ell}{\partial \sigma_\mathcal{B}^2}$ can be re-written as:
\begin{equation}
\begin{split}
\frac{\partial \ell}{\partial \mu_\mathcal{B}} & = 
\frac{\partial \ell}{\partial \mu_\mathcal{JS}} \cdot \frac{\partial \mu_\mathcal{JS}}{\partial \mu_\mathcal{B}} +
\frac{\partial \ell}{\partial \mathcal{S}_{\mu_\mathcal{B}}} \cdot \frac{\partial \mathcal{S}_{\mu_\mathcal{B}}}{\partial \mu_\mathcal{B}} +
\frac{\partial \ell}{\partial \sigma_{\mu_\mathcal{B}}^2} \cdot \frac{\partial \sigma_{\mu_\mathcal{B}}^2}{\partial \mu_\mathcal{B}}
\end{split}
\end{equation}
\begin{equation}
\begin{split}
\frac{\partial \ell}{\partial \sigma_\mathcal{B}^2} & = 
\frac{\partial \ell}{\partial \sigma_\mathcal{JS}^2} \cdot \frac{\partial \sigma_\mathcal{JS}^2}{\partial \sigma_\mathcal{B}^2} +
\frac{\partial \ell}{\partial \mathcal{S}_{\sigma_\mathcal{B}^2}} \cdot \frac{\partial \mathcal{S}_{\sigma_\mathcal{B}^2}}{\partial \sigma_\mathcal{B}^2} +
\frac{\partial \ell}{\partial \sigma_{\sigma_\mathcal{B}^2}^2} \cdot \frac{\partial \sigma_{\sigma_\mathcal{B}^2}^2}{\partial \sigma_\mathcal{B}^2}
\end{split}
\end{equation}

\end{document}